\documentclass{article}

\newif\ifcomment\commentfalse
\usepackage[a-1b]{pdfx}

\usepackage{framed}
\usepackage{mdwlist}
\usepackage{siunitx}
\usepackage{latexsym}
\usepackage{colortbl}
\usepackage{xcolor}
\usepackage{nicefrac}
\usepackage{booktabs}
\usepackage{fnpct}
\usepackage{amsfonts}
\usepackage[T1]{fontenc}
\usepackage{bold-extra}
\usepackage{amsmath}
\usepackage{amssymb}
\usepackage{bm}
\usepackage{graphicx}
\usepackage{mathtools}
\usepackage{microtype}
\usepackage{multirow}
\usepackage{multicol}
\usepackage{xpatch}
\usepackage{latexsym,comment}
\usepackage[normalem]{ulem}

\newcommand*{\missingreference}{{\Huge \colorbox{red}{?reference?}}}
\newcommand*{\missingcitation}{{\Huge \colorbox{red}{?citation?}}}

\makeatletter
\xpatchcmd{\@setref}{\bfseries}{\missingreference}{}{}
\def\@citex[#1]#2{\leavevmode
    \let\@citea\@empty
    \@cite{\@for\@citeb:=#2\do
        {\@citea\def\@citea{,\penalty\@m\ }%
            \edef\@citeb{\expandafter\@firstofone\@citeb\@empty}%
            \if@filesw\immediate\write\@auxout{\string\citation{\@citeb}}\fi
            \@ifundefined{b@\@citeb}{\hbox{\reset@font\missingcitation}%
                \G@refundefinedtrue
                \@latex@warning
                {Citation `\@citeb' on page \thepage \space undefined}}%
            {\@cite@ofmt{\csname b@\@citeb\endcsname}}}}{#1}}
\makeatother

\newcommand{\gem}[1]{\mbox{\textsc{gem}}}

\newcommand{\hidetext}[1]{}
\newcommand{\ignore}[1]{}

\ifcomment
    \newcommand{\pinaforecomment}[3]{\colorbox{#1}{\parbox{.8\linewidth}{#2: #3}}}

    \newcommand{\prtodo}[1]{\pinaforecomment{lightblue}{pr}{#1}}
    \newcommand{\prtodoi}[1]{\pinaforecomment{lightblue}{pr}{#1}}
\else
    \newcommand{\pinaforecomment}[3]{}
    \newcommand{\prtodo}[1]{}
    \newcommand{\prtodoi}[1]{}
\fi

\newcommand{\smallurl}[1]{ \begin{tiny}\url{#1}\end{tiny}}

\definecolor{lightblue}{HTML}{3cc7ea}
\definecolor{CUgold}{HTML}{CFB87C}
\definecolor{grey}{rgb}{0.95,0.95,0.95}
\definecolor{ceil}{rgb}{0.57, 0.63, 0.81}
\definecolor{UMDred}{HTML}{ed1c24}
\definecolor{UMDyellow}{HTML}{ffc20e}

\usepackage[position, preprint]{neurips_2026}

\usepackage[utf8]{inputenc} % allow utf-8 input
\usepackage[T1]{fontenc}    % use 8-bit T1 fonts
\usepackage{hyperref}       % hyperlinks
\usepackage{url}            % simple URL typesetting
\usepackage{booktabs}       % professional-quality tables
\usepackage{amsfonts}       % blackboard math symbols
\usepackage{nicefrac}       % compact symbols for 1/2, etc.
\usepackage{microtype}      % microtypography
\usepackage{xcolor}         % colors

\usepackage{booktabs}
\usepackage{longtable}
\usepackage{xcolor}
\usepackage{array}
\usepackage{geometry}
\usepackage{caption}
\usepackage{svg}
\usepackage{booktabs}
\usepackage{tabularx}
\definecolor{eduFill}{RGB}{250,238,218}
\definecolor{eduStroke}{RGB}{186,117,23}
\definecolor{eduText}{RGB}{99,56,6}
 
\definecolor{swFill}{RGB}{230,241,251}
\definecolor{swStroke}{RGB}{24,95,165}
\definecolor{swText}{RGB}{12,68,124}
 
\definecolor{hcFill}{RGB}{250,236,231}
\definecolor{hcStroke}{RGB}{153,60,29}
\definecolor{hcText}{RGB}{113,43,19}
 
\definecolor{grayFill}{RGB}{241,239,232}
\definecolor{grayStroke}{RGB}{95,94,90}
\definecolor{grayText}{RGB}{68,68,65}

 \definecolor{lawFill}{RGB}{215, 200, 235}    % muted violet
\definecolor{lawStroke}{RGB}{120,80,200}
\definecolor{lawText}{RGB}{70,40,140}

\definecolor{rowgray}{HTML}{F7F7F7}

\title{Benchmarked Yet Not Measured - Generative AI Should be Evaluated Against Real-World Utility}

\author{%
  Ishani Mondal, Shweta Bhardwaj \\
        University of Maryland, College Park, USA  \\
}

\begin{document}

% \usetikzlibrary{positioning, arrows.meta, shapes.geometric, fit, backgrounds}
\maketitle

\begin{abstract}

Generative AI systems achieve impressive performance on standard benchmarks yet fail to deliver real-world utility—a disconnect we found across 28 deployment cases spanning education, healthcare, software engineering, and law. 
We argue that this benchmark--utility gap arises from three recurring failures: \textbf{proxy displacement}, \textbf{temporal collapse} and \textbf{distributional concealment}, reinforced by Goodhart-style optimization on saturated benchmarks and a broader failure of construct validity in current evaluation practice.
\textbf{Motivated by these, we argue that generative AI evaluation requires a paradigm shift: from static benchmark-centered transparency toward stakeholders, goals, and context-conditioned utility transparency grounded in human outcome trajectories.} Existing evaluations primarily characterize properties of model outputs, while deployment success depends on whether interaction with AI improves stakeholders’ ability to achieve their goals over time. The missing construct is therefore \emph{utility}: the change in a stakeholder’s capability induced through sustained interaction with an AI system within a deployment context.
To operationalize this perspective, we propose \textbf{SCU-GenEval}, a four-stage evaluation framework consisting of stakeholder--goal mapping, construct--indicator specification, mechanism modeling, and longitudinal utility measurement. To make these stages practically deployable, we introduce three supporting instruments: structured deployment protocols, context-conditioned user simulators, and persona--goal-conditioned proxy metrics. We conclude with domain-specific calls to action, arguing that progress in generative AI must be evaluated by measurable improvements in human outcomes rather than by benchmark performance alone.
\end{abstract}

% intro
\section{Introduction}

Benchmarks such as ImageNet~\citep{Russakovsky2015}, GLUE~\citep{wang-etal-2018-glue}, HumanEval~\citep{DBLP:journals/corr/abs-2107-03374}, and VBench~\citep{huang2023vbench} have been central to the progress of generative AI by enabling standardized comparison across models. However, as generative AI systems move into education, healthcare, software engineering, and other high-stakes settings, a growing disconnect has emerged between benchmark performance and real-world usefulness. Current evaluation pipelines largely focus on artifact-level metrics such as BLEU \citep{papineni-etal-2002-bleu}, ROUGE \citep{lin-2004-rouge}, pass@k \citep{DBLP:journals/corr/abs-2107-03374}, fluency, preference win-rates, and visual realism. In deployment, however, the critical question is fundamentally different: \textbf{does interaction with the system improve human capability, decision-making, learning, collaboration, and trust over time in real environments?}

\begin{figure}[t]
\centering

% ---------------- Figure ----------------
\includegraphics[width=\linewidth]{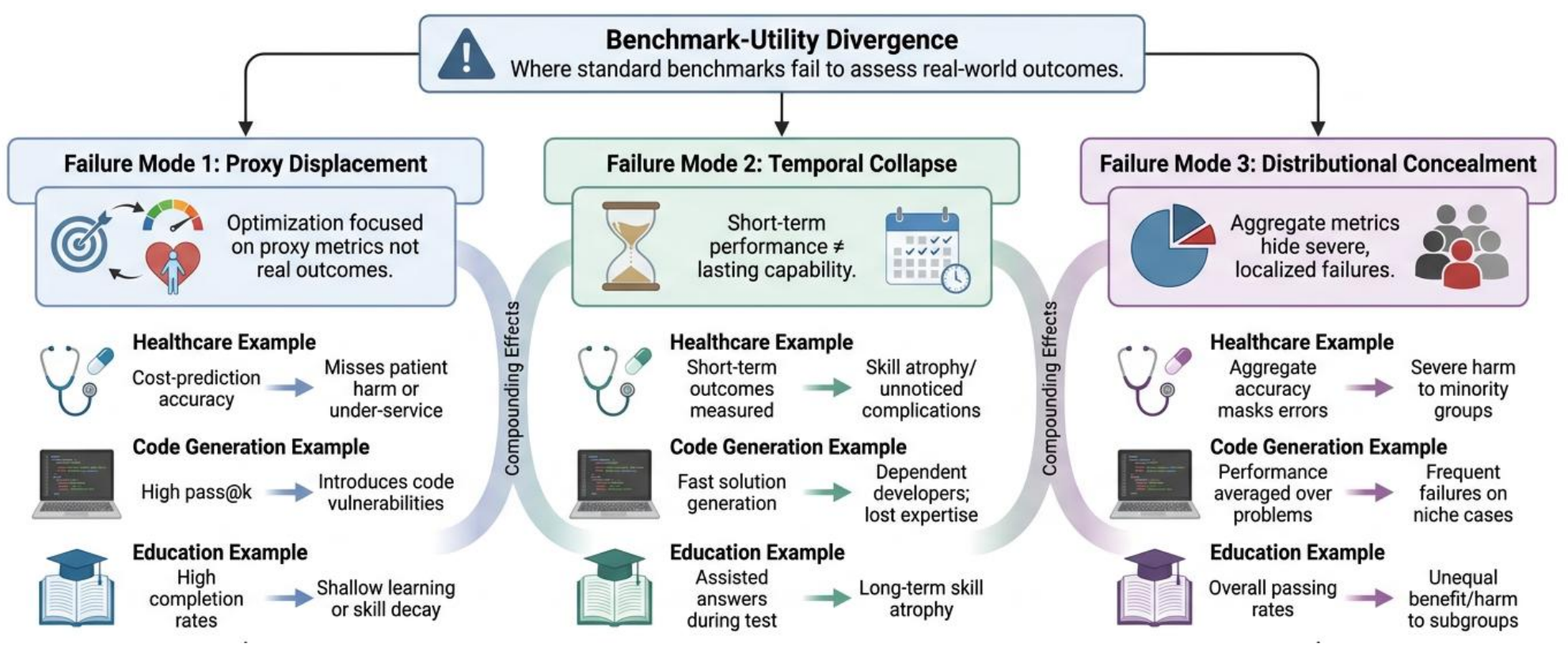}
\vspace{1mm}

\scriptsize
\renewcommand{\arraystretch}{1.12}
\setlength{\tabcolsep}{3pt}

\begin{tabular}{@{} p{0.22\linewidth} p{0.73\linewidth} @{}}
\toprule
\textbf{Failure Mode} & \textbf{Case IDs and Description} \\
\midrule

\textbf{Proxy displacement} &
S6 (fabricated citations, fluency), S10 (Copilot $\sim$40\% vulnerabilities), S13 (coverage overstates fault-detection), S15 (BLEU misses doc errors), S16 (cost $\neq$ health need), S22 (medical LLM hallucinations), S25 (COMPAS AUC $\neq$ contestable justice), S26/S28 (fluent legal citations $\neq$ valid authority) \\

\addlinespace[1mm]

\textbf{Temporal collapse} &
S1 (ITS transfer unmeasured), S3 (ChatGPT completion vs.\ transfer), S4 (AI summaries skip retrieval), S5 (CS1 Copilot metacognitive harm), S14 (developer skill decay), S24 (chatbot short-term-only gains) \\

\addlinespace[1mm]

\textbf{Distributional concealment} &
S7 (low-performers gain less), S16 (Black patients under-enrolled), S17 (COVID classifier confounders), S18 (GPT-4 racial/gender bias), S20 (Watson site-generalization failure), S27 (SafeRent score obscured voucher/race disparities) \\

\bottomrule
\end{tabular}

\vspace{-2mm}

\caption{
\textbf{Benchmark--utility divergence in generative AI evaluation.}
The figure illustrates three recurring structural failures of current evaluation: \emph{proxy displacement}, \emph{temporal collapse}, and \emph{distributional concealment}. The table below preserves full set of representative cases for each failure.
}
\label{fig:benchmark_utility_divergence}

\end{figure}

To investigate this benchmark--utility disconnect, we analyze 28 documented deployment cases spanning education, healthcare, and software engineering (Figure~\ref{fig:benchmark_utility_divergence}; detailed in Table~\ref{tab:cases}). Across domains, we observe a recurring pattern: systems that perform strongly on standard benchmarks often fail to deliver reliable real-world utility. In software engineering, code-generation systems with high pass@k scores still introduce security vulnerabilities in nearly 40\% of generated solutions~\citep{Asleep}. In healthcare, the widely deployed Epic Sepsis model missed 67\% of sepsis cases while incorrectly alerting on 18\% of patients despite strong internal validation~\citep{Wong2021-fi, Habib-EpicSepsis2021}. In education, systems that improve short-term task completion frequently fail to improve long-term retention or transfer, while AI-assisted programming tools can reduce metacognitive engagement and contribute to skill atrophy over time. Collectively, these findings suggest that benchmark success is often a weak proxy for the human outcomes deployment seeks to improve.

% \begin{table}[!t]
% \small
% \centering
% \renewcommand{\arraystretch}{1.3}
% \begin{tabular}{@{} p{0.18\textwidth} p{0.75\textwidth} @{}}
% \toprule
% \textbf{Failure Mode} & \textbf{Case IDs and Description} \\
% \midrule
% Proxy displacement &
% Tractable metrics treated as valid without construct validity: S6 (fabricated citations, fluency), S10 (Copilot $\sim$40\% vulnerabilities), S13 (coverage overstates fault-detection), S15 (BLEU misses doc errors), S16 (cost $\neq$ health need), S22 (medical LLM hallucinations), S25 (COMPAS AUC $\neq$ contestable justice), S26/S28 (fluent legal citations $\neq$ valid authority) \\
% \addlinespace
% Temporal collapse &
% Snapshot/assisted performance mistaken for durable capability: S1 (ITS transfer unmeasured), S3 (ChatGPT completion vs.\ transfer), S4 (AI summaries skip retrieval), S5 (CS1 Copilot metacognitive harm), S14 (developer skill decay), S24 (chatbot short-term-only gains) \\
% \addlinespace
% Distributional concealment &
% Aggregate scores hide subgroup/site harms: S7 (low-performers gain less), S16 (Black patients under-enrolled), S17 (COVID classifier confounders), S18 (GPT-4 racial/gender bias), S20 (Watson site-generalization failure), S27 (SafeRent score obscured voucher/race disparities) \\
% \bottomrule
% \end{tabular}
% \caption{Structural failure modes of current generative AI evaluation.}
% \label{tab:failure-modes}
% \end{table}

From these cases, we identify three recurring structural failures underlying the benchmark--utility gap. 
First, \textbf{proxy displacement} occurs when tractable benchmark metrics are treated as valid measures of complex real-world constructs without evidence of construct validity. Metrics such as BLEU, ROUGE, pass@k, and preference win-rates are routinely interpreted as indicators of semantic fidelity, code quality, or usefulness despite weak alignment with deployment outcomes \citep{Sai}. This also manifests when evaluations focus only on outcomes while ignoring whether systems follow acceptable reasoning processes: transparency, contestability, or domain-specific procedural values specially in high-stakes domains \citep{linna-aiFailsLegalReasoning2026}.
Second, \textbf{temporal collapse} occurs when short-term or AI-assisted performance is mistaken for durable capability. Existing evaluations frequently measure immediate task success while ignoring retention, transfer, over-reliance, metacognitive decline, calibration, and long-term skill development \cite{RODRIGUES2026106252, livebench}. 
Third, \textbf{distributional concealment} manifests when average scores mask the failures across populations, under-represented groups, or unseen deployment settings, allowing systems that appear robust on average to systematically fail for minority populations, novice users, or unseen environments \citep{ZHANG2026101515}.
% Fourth, \textbf{normative and procedural invisibility} occurs when evaluations focus only on outcomes while ignoring whether systems follow acceptable reasoning processes: transparency, contestability, or domain-specific procedural values specially in high-stakes domains \citep{linna-aiFailsLegalReasoning2026}.

Importantly, these failures are not isolated. They reinforce one another through benchmark saturation, contamination, memorization, and Goodhart-style optimization dynamics~\citep{Goodhart1985, akhtar2026aibenchmarksplateausystematic, jacovi-etal-2023-stop, huang-etal-2022-large, carlini2019secretsharerevaluatingtesting}. More fundamentally, they expose a broader problem of construct validity~\citep{Cronbach1955-gd, Jacobs-constructValidity2021}: current benchmarks are rarely validated against the deployment outcomes they are implicitly assumed to represent~\citep{McIntosh_2026}. Prior work has already questioned whether automatic metrics meaningfully capture human utility~\citep{schmidtova-etal-2024-automatic-metrics, roy2026prototypicalitybiasrevealsblindspots, xu-etal-2023-instructscore}, while research in human--AI interaction demonstrates that usefulness varies substantially across users, goals, and contexts~\citep{GONDOCS2025103622}. Together, these findings suggest that aggregate benchmark scores provide only partial transparency about deployed AI systems.

% A deeper pattern across many documented failures is that current evaluations systematically prioritize \emph{immediate assisted performance} over \emph{durable independent capability}. Systems often appear successful because they reduce cognitive effort, accelerate completion, or automate difficult reasoning steps. Yet educational and cognitive science research consistently shows that productive struggle, effortful retrieval, verification, and active reasoning are central mechanisms underlying long-term learning and expertise development~\citep{Roediger2006-xp, Retrieval}. As a result, a system can improve benchmark performance while simultaneously weakening the very human capabilities deployment intends to support. Higher assisted accuracy, lower friction, or faster completion should therefore not automatically be interpreted as positive outcomes. In many real-world settings, the central question is not whether AI removes effort, but whether it improves human capability without degrading retention, transfer, calibration, collaboration, or independent reasoning over time.

\textbf{\emph{Position.}} These observations motivate the central position of this paper: \textbf{generative AI evaluation requires a paradigm shift from static benchmark-centered transparency toward stakeholder-conditioned, goal-conditioned, and task-conditioned utility transparency centered on human outcome trajectories and capability shifts in specific deployment context}. Existing benchmarks primarily provide transparency about \emph{model outputs}; deployment instead requires transparency about \emph{how interaction with AI changes human capability over time}. Under this view, evaluation should not merely report whether a model produces high-quality outputs in isolation, but instead characterize whose capabilities change, under which goals and deployment contexts, through which mechanisms, and across what temporal horizons. In other words, the unit of evaluation --- and therefore the unit of transparency --- must shift from \emph{artifact quality} to the \emph{human outcome trajectory reflecting their change in capability}.

\textbf{\emph{Our Proposal.}} Based on these observations, we argue that the missing concept in current evaluation is \emph{utility}~\citep{ethayarajh-jurafsky-2020-utility, linzen-2020-accelerate}: not simply the quality of an output artifact, but the extent to which interaction with an AI system changes a person’s ability to achieve their goals over time. To address this gap, we propose \textbf{Stakeholder-Centered Utility Evaluation (SCU-GenEval)}, a framework for evaluating generative AI systems through deployment-grounded human outcomes. The framework operationalizes utility through four connected stages: identifying stakeholders and deployment goals, defining constructs and measurable indicators, modeling the mechanisms through which AI changes user behavior and capability, and finally measuring utility longitudinally and across subgroups. \emph{In this sense, SCU-GenEval is not merely a new benchmark methodology, but a broader proposal for a measurement science centered on stakeholder-conditioned utility transparency rather than benchmark performance alone}. To operationalize this framework, we introduce three complementary instruments.
% More broadly, we call for a shift in how progress in generative AI is measured. Evaluation must move beyond static artifact scoring toward deployment-grounded measurement practices that capture transfer, robustness, subgroup effects, longitudinal capability, and procedural validity. We encourage benchmark maintainers, model developers, regulators, and funding agencies to adopt domain-specific utility reporting standards, such as unassisted transfer outcomes in education, externally validated and demographically disaggregated evaluation in healthcare, and long-term developer capability measures in software engineering. 
We further argue that the field needs shared deployment datasets, reporting protocols, and evaluation standards analogous to CONSORT (Consolidated Standards of Reporting Trials)\footnote{\url{https://en.wikipedia.org/wiki/Consolidated_Standards_of_Reporting_Trials}}
 and TRIPOD\footnote{\url{https://www.equator-network.org/reporting-guidelines/tripod-statement/}}.

\textbf{\emph{Our Contributions.}} This paper makes four contributions. First, we analyze 28 deployment cases (Table~\ref{tab:cases}) showing that the benchmark--utility gap is systematic, cross-domain, and recurring rather than anecdotal. 
Second, we identify three structural causes underlying this gap: proxy displacement, temporal collapse and distributional concealment (Section~\ref{sec:failure-modes}). Third, we introduce SCU-GenEval, a stakeholder-centered framework for measuring utility as changes in human capability over time (Section~\ref{sec:proposal}). 
Finally, we provide concrete recommendations for researchers, benchmark creators, industry practitioners, and policymakers to develop evaluation practices that measure real-world usefulness rather than benchmark performance alone (Section~\ref{sec:calls-to-action}).
% =====================================================================
%  Section 2 rewrite: "Evidence: Where the Benchmark-Utility Gap
%  Manifests and What It Costs"
%
%  This file is standalone-compilable in Overleaf (just hit Recompile).
%  To use in your main paper:
%    1. Copy the package imports below into your preamble (skip any
%       you already have).
%    2. Copy the color definitions into your preamble.
%    3. Copy everything between the BEGIN PAPER CONTENT / END PAPER
%       CONTENT markers into your paper, replacing the old Section 2.

\section{Where Does the Benchmark--Utility Gap Manifest and What It Costs?}
\label{sec:failure-modes}
% \label{sec:evidence}
\vspace{-2.2mm}
\vspace{-1.2mm}
Evaluation pipelines fail along three reinforcing axes—\textit{what} they measure, \textit{when} they measure, and \textit{how} they aggregate—yielding scores that are internally consistent but externally invalid. The 25 cases in Table~\ref{tab:cases} instantiate these failures, compounded by contamination, LLM-as-judge circularity, and the lack of interaction-grounded proxies.
\paragraph{Proxy Displacement: Tractable Metrics Are Not Valid Measures.}
\label{sec:proxy-displacement}
\vspace{-1.2mm}
\textit{Without construct validity, benchmark gains are not capability gains.}
BLEU \citep{papineni-etal-2002-bleu}, ROUGE, pass@k \citep{DBLP:journals/corr/abs-2107-03374}, and AUC are routinely treated as proxies for quality or capability without validation \citep{Cronbach1955-gd, Jacobs_2021}. Systems optimize the proxy while regressing on deployment-relevant properties—an effect amplified by data contamination and LLM-as-judge circularity, and predicted by Goodhart's law \citep{Goodhart1985}.
Besides, reasoning traces can diverge from the computation that produced an output—recently documented in faithfulness work \citep{Turpin-reasoning-neurips2023}—hence, stated justifications cannot serve as evidence that procedural standards are met. In adjudicative and clinical settings, where transparency and contestability are legally required \citep{linna-aiFailsLegalReasoning2026, Kiseleva-Healthcare-transparency2022}, output-based evaluation cannot confirm compliance.
\paragraph{Temporal Collapse: Snapshot Performance Is Not Durable Capability.}
\label{sec:temporal-collapse}
\vspace{-1.2mm}
\textit{Assisted, in-the-moment accuracy says little about retention, transfer, or unassisted competence.}
Pass@k, in-system accuracy, and completion rates capture point-in-time performance but ignore retention and transfer—contradicting evidence that durable learning requires effortful retrieval \citep{Roediger2006-xp, Retrieval} and that automation induces skill decay \citep{BAINBRIDGE1983775, Parasuraman2010-gq}. Static, non-interactive benchmarks cannot track how capability evolves under feedback, systematically overstating utility.
\paragraph{Distributional Concealment: Aggregates Hide the Harms That Matter.}
\label{sec:distributional-concealment}
\vspace{-1.2mm}
\textit{High averages coexist with severe, localized failure.}
Average accuracy and AUC can stay high while specific populations or shifted distributions fail badly—well documented in fairness and robustness work \citep{pmlr-v81-buolamwini18a, Koenecke} and the basis for disaggregated reporting \citep{Mitchell2018, gebru}. Dataset-bound evaluation misses real-world variability and interaction context.

% \paragraph{Procedural Invisibility: Outputs Alone Cannot Verify Procedural Norms}
% \textit{A model's stated reasoning is not evidence that domain-specific procedural norms are met.} Reasoning traces can diverge from the computation that produced an output—recently documented in faithfulness work \citep{Turpin-reasoning-neurips2023}—hence, stated justifications cannot serve as evidence that procedural standards are met. In adjudicative and clinical settings, where transparency and contestability are legally required \citep{linna-aiFailsLegalReasoning2026, Kiseleva-Healthcare-transparency2022}, output-based evaluation cannot confirm compliance.

\textit{The failures reinforce rather than cancel; contemporary pipelines make the compounded effect worse.}
Misaligned proxies, snapshot overstatement, and aggregate concealment interact—and contaminated data, circular judges, and non-interactive tasks magnify each. The result is evaluation that is internally consistent but externally invalid.

\section{Proposal: Stakeholder-Centered Utility Evaluation (SCU-GenEval)}
\label{sec:proposal}
% \vspace{-2.2mm}
% We propose \textbf{SCU-GenEval}, a framework for evaluating generative AI by what it does for people, not by what it produces. The framework has two parts. The first part is a pipeline of four stages that an evaluator works through in order. The second part is a set of three instruments that make the stages practical to run. Together, the stages and instruments are designed to address the three failure modes from Section~\ref{sec:failure-modes}: \textbf{proxy displacement}, \textbf{temporal collapse}, and \textbf{distributional concealment}. We point out which component fixes which failure as we go.

\begin{figure}[t]
\centering
\includegraphics[width=\linewidth]{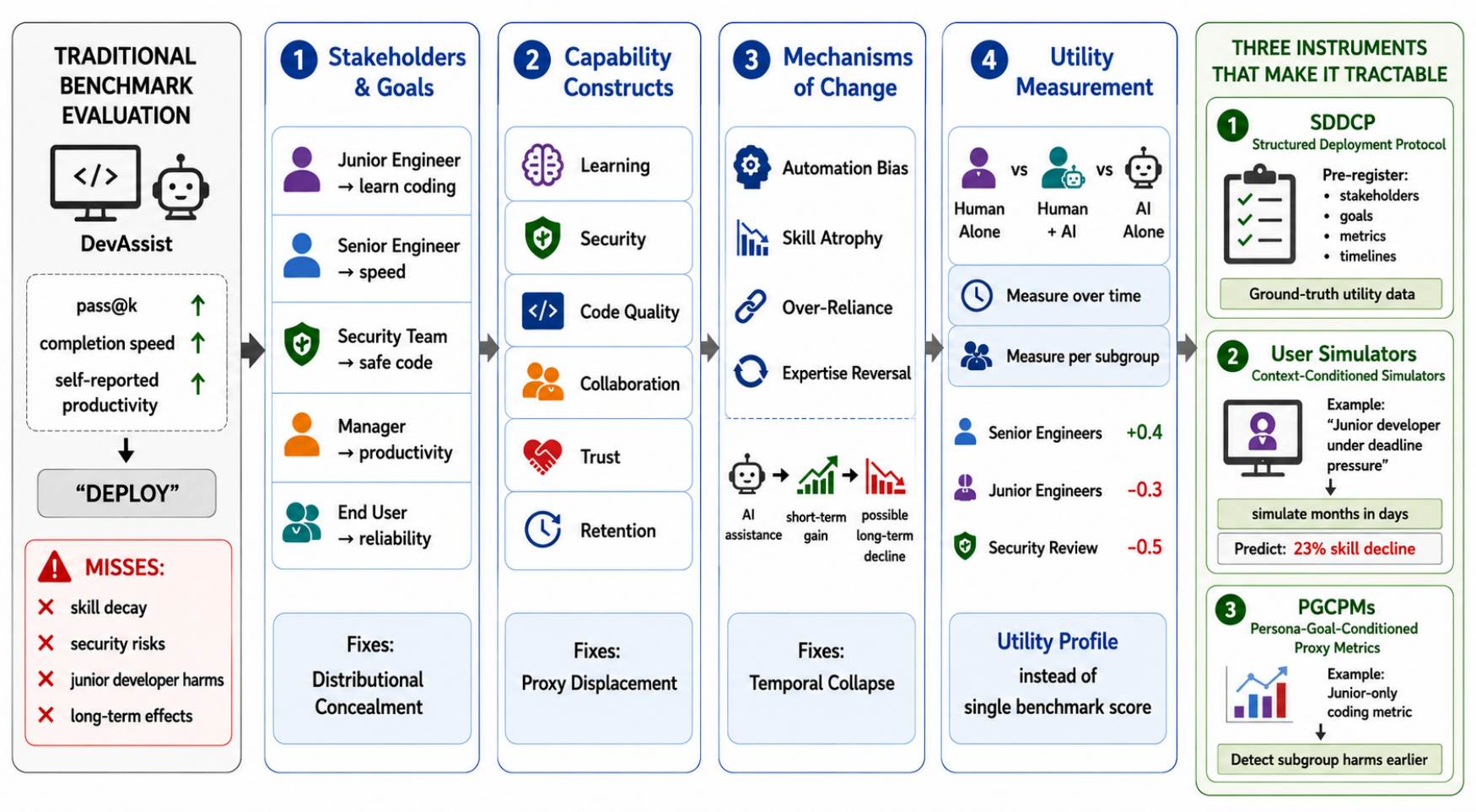}
\caption{SCU-GenEval applied to a concrete yet hypothetical DevAssist rollout scenario. \textbf{Left:} A standard benchmark tunnel sees only pass@k, self-reported productivity and outputs ``Deploy.'' \textbf{Middle:} the four SCU-GenEval stages enumerate stakeholders, name constructs, identify mechanisms, and measure utility per stakeholder. \textbf{Right:} Three instruments make the framework practical.}
\label{fig:scu-walkthrough}
\vspace{-2.2mm}
\end{figure}
\vspace{-2.2mm}

Section~\ref{sec:failure-modes} documented the structural failures of current evaluation. Before proposing a framework, we make explicit what these failures share, because the shared structure determines what any solution \textit{must do}. This diagnosis yields the following requirements that any evaluation framework must satisfy:
\begin{itemize} \item \textbf{(R1) User-Outcome centered measurement.} The unit of evaluation must be the user's outcome (capability, decision, retention) as a result of using the AI, not properties of the AI's output. Evaluation must also check the system's reasoning process against domain-specific procedural standards through external audit rather than the system's own stated explanations. \item \textbf{(R2) Temporal measurement.} Evaluation must measure change over time, including unassisted performance \emph{after} AI exposure, not single-shot performance \emph{during} AI exposure. \item \textbf{(R3) Disaggregated measurement.} Evaluation must report outcomes per (user, goal, context) tuple, not averaged across heterogeneous users. 
% \item\textbf{(R4) Procedural verification}. Evaluation must check the system's reasoning process against domain-specific procedural standards through external audit rather than the system's own stated explanations. This is the structural fix for procedural invisibility.
\end{itemize} We argue that the missing anchor unifying these requirements is \textbf{utility}~\citep{ethayarajh-jurafsky-2020-utility, linzen-2020-accelerate}: the change in a user's ability to achieve their goal because they used the AI. Utility is user-centered (R1), inherently temporal (R2), and necessarily indexed by user and context (R3).  
% \textcolor{red}{However, in high-stakes domains, even high-utility systems may bypass procedural standards that legitimacy requires (R4). Our framework first formalizes utility (\S\ref{subsec:quantifying-utility}), audits procedural adequacy as a separate deployment gate (add reference to audit section here), and proposes a theoretical framework around them} (\S\ref{subsec:framework}, S\ref{subsec:instruments}).
The remainder of this section formalizes utility (\S\ref{subsec:quantifying-utility}) and builds a framework around it (\S\ref{subsec:framework} and \S\ref{subsec:instruments}).

% \S\ref{subsec:fits-together}).

\subsection{Quantifying Utility} \label{subsec:quantifying-utility} 

Inspired by \citep{CAMPAGNER2022106930,ethayarajh-jurafsky-2020-utility}, we quantify utility as the counterfactual change in a stakeholder’s capability to achieve a goal through interaction with an AI system under a particular deployment context. 
%Our formulation is grounded in the capability approach to human welfare [Sen, 1999, Nussbaum, 2011], which treats well-being as a person's substantive freedom to do and be what they have reason to value; we develop the full philosophical foundation in Appendix X.
Capability is treated as a vector of indicators rather than a scalar, and the temporal index $t$ distinguishes performance-\emph{with}-AI assistance from performance-\emph{without}-it after
exposure.
Let $\mathbf{C}(u, G, D, t) \in \mathbb{R}^{K}$ denote the capability vector of
stakeholder $u$ pursuing a goal $G$ in deployment context $D$ at the horizon $t$,
with components:
\begin{equation}
    \mathbf{C}(u, G, D, t) \;=\; \bigl(I_1(u,G,D,t),\; I_2(u,G,D,t),\; \ldots,\; I_K(u,G,D,t)\bigr),
\end{equation}
where each $I_k$ is a measurable indicator of capability (e.g., decision
quality, retention, transfer, calibration, safety, or collaboration effectiveness). Utility is then the difference between capability after and before
interaction with the system:
\begin{equation}
    \mathbf{U}(u, G, y, D, t)
    \;=\;
    \mathbf{C}_{\text{after}}(u, G, y, D, t)
    \;-\;
    \mathbf{C}_{\text{before}}(u, G, D, t),
    \label{eq:utility}
\end{equation}

% \begin{equation}
% U(u,G,y,D,t)
% =
% C_{\text{after}}(u,G,y,D,t)
% \end{equation}
% \begin{equation}
% C_{\text{before}}(u,G,D,t),
% \quad
% C(u,G,D,t)
% =
% \sum_{k=1}^{K}
% w_k(u,G,D)\cdot I_k(u,G,D,t)
% \end{equation}

% \begin{equation}
% U(u,G,y,D,t)
% =
% C_{\text{after}}(u,G,y,D,t)
% \end{equation}
% \begin{equation}
% C_{\text{before}}(u,G,D,t),
% \quad
% C(u,G,D,t)
% =
% \sum_{k=1}^{K}
% w_k(u,G,D)\cdot I_k(u,G,D,t)
% \end{equation}

where $y$ denotes the AI system and its outputs. The reference state
$\mathbf{C}_{\text{before}}$ may be either the user's pre-exposure baseline
or the user's unassisted capability at the same horizon, depending on
whether the evaluation targets durable change or in-the-moment assistance.
Deployment decisions ultimately require an ordering, for which a scalar
summary $\tilde{U} = \sum_k w_k(u, G, D)\, U_k$ may be computed with
stakeholder- and context-specific weights $w_k$; we treat this as a
reporting step rather than a definitional one: aggregation is itself a form of concealment.
% If a scalar summary is required downstream, it can be computed as $\tilde{U} = \sum_k w_k(u,G,D)\, U_k$ with stakeholder- and context-specific weights $w_k$; we treat this as a reporting step rather than a definitional one, since aggregation is the mechanism behind distributional concealment.

% where \(u\) denotes the stakeholder or user population, \(G\) the stakeholder-conditioned goal, \(y\) the AI system and its outputs, \(D\) the deployment context, and \(t\) the temporal horizon at which capability is measured. \(I_k\) denotes measurable indicators of capability (e.g., decision quality, retention, transfer, calibration, safety, or collaboration effectiveness), while \(w_k\) denotes stakeholder- and context-specific importance weights.

This formulation operationalizes the requirements specified above. 
\textbf{First, utility is stakeholder-centered and deployment-conditioned
(R1):} defining $\mathbf{U}$ as change in $\mathbf{C}$ rather than a property of generated artifact $y$ is the structural fix for proxy displacement. Since
stakeholder capability is conditioned on $D$, the same system may produce different utility profiles across expertise levels, workflows, or institutional constraints, and utility may even become negative
when interaction induces overreliance, unsafe behavior, inequity, or long-term skill atrophy.\\
\textbf{Second, utility is temporal (R2):} the choice of reference state for $\mathbf{C}_{\text{before}}$ at horizon $t$ separates short-term assistance
(unassisted same-horizon reference) from durable capability change (pre-exposure reference), distinguishing transient performance amplification
from sustained learning or transfer.
\textbf{Finally, utility is indexed and disaggregated (R3):} keeping
$\mathbf{U}$ as a vector over stakeholder–goal–context tuples $(u, G, D)$ rather than collapsing to an abstract ``average user'' or single score is the structural fix for distributional concealment, since aggregate scores can hide severe subgroup harms despite strong overall performance.

\subsection{The SCU-GenEval Framework: Operationalizing Utility}
\label{subsec:framework}
\vspace{-1.2mm}

 To demonstrate both the \emph{measurement target} (\textit{what}) and the \emph{measurement process} (\textit{how}), Figure~\ref{fig:scu-walkthrough} walks through a concrete yet hypothetical deployment decision: a large organization evaluating whether to roll out an AI coding assistant (``DevAssist'') to 5,000 engineers. 
 The figure is designed to be read from left to right as a progression from conventional evaluation toward stakeholder-centered utility evaluation. 
 The leftmost region introduces the deployment scenario and the standard evaluation pipeline. 
 Here, the vendor reports familiar benchmark evidence --- 89\% pass@k and 32\% faster ticket completion --- which then flows through a narrow ``Standard Benchmark Tunnel'' consisting of pass@k, self-reported productivity, and win-rate comparisons against the existing tool. 
 Because this pipeline evaluates only artifact-level performance proxies, it collapses heterogeneous outcomes into a single recommendation: ``Deploy.''

The central region of the figure expands this narrow pipeline into the four-stage SCU-GenEval process. Each stage incrementally refines the evaluation from output quality toward longitudinal human utility. Stage~1 identifies \emph{who} is affected and \emph{what goals} they are pursuing. Stage~2 formalizes \emph{what capability changes actually matter} for those users. Stage~3 models \emph{how interaction with the system changes those capabilities over time}. Stage~4 then operationalizes these assumptions into direct utility measurement. The stages are intentionally sequential: each stage produces the inputs required by the next. Stakeholders and goals identified in Stage~1 determine which constructs are formalized in Stage~2; the constructs from Stage~2 determine which mechanisms of change must be modeled in Stage~3; and the mechanisms identified in Stage~3 determine what longitudinal measurements Stage~4 must perform. In this way, the framework forms a connected causal chain rather than a collection of independent evaluation modules.

The rightmost panel of Figure~\ref{fig:scu-walkthrough} introduces the three supporting instruments --- User Simulators, Structured Deployment Data Collection Protocols (SDDCP), and Persona-Goal-Conditioned Proxy Metrics (PGCPMs) --- that make the four-stage process practical to execute at deployment scale.
%A callout at the top of the figure previews the central finding: although the standard benchmark pipeline produced an overall positive deployment recommendation, SCU-GenEval reveals sharply different utility profiles across groups, yielding $(+0.4, -0.3, -0.5)$ utility for senior engineers, junior engineers, and security review teams respectively. The figure therefore visually illustrates the core argument of the paper: aggregate benchmark success can conceal subgroup-specific harms that only become visible once evaluation is grounded in longitudinal human utility. A legend at the bottom right links each visual component to the three structural failure modes introduced in Section~\ref{sec:failure-modes}, and we refer back to specific panels throughout the section so that each abstract concept remains tied to a concrete deployment scenario.

\paragraph{Four Stages: From Definition to Measurement.}
As explained above, the four stages together operationalize the utility formulation.
Rather than functioning as isolated procedures, each stage constrains and informs the next, transforming abstract notions of utility into measurable quantities. 
Collectively, they specify the user $u$, the goal $G$, the capability change $\mathbf{C}_{\text{after}} - \mathbf{C}_{\text{before}}$, the temporal dynamics governing that change, and finally the empirical protocols required to estimate utility.

\textbf{Stage 1: Stakeholders and Goals --- specifying $u$ and $G$.}
The first stage identifies the users $u$ affected by deployment and the goals $G$ they are attempting to achieve. The output is a \textbf{Stakeholder--Goal Map} that explicitly enumerates whose outcomes matter and what successful assistance means for each group. This stage establishes the foundation for the remainder of the framework: before utility can be measured, the framework must first define \emph{for whom} utility is being measured and \emph{toward what objective}. Naming stakeholders and goals up front directly counters distributional concealment by preventing heterogeneous populations from being silently merged into a single average, while also providing the first safeguard against proxy displacement by ensuring that the intended human outcome is specified before proxy metrics are introduced. 
Several failures from Section~\ref{sec:failure-modes} originate precisely from the absence of this step, including S16, where optimization targeted insurer cost rather than patient need, and S20, where the relevant clinician populations were never clearly specified. In Figure~\ref{fig:scu-walkthrough}, the Stage~1 panel distinguishes five DevAssist stakeholder groups --- senior engineers, junior engineers, security engineers, engineering managers, and end users --- each associated with distinct proximate and distal goals. By contrast, the standard benchmark tunnel collapses all of them into the single abstraction of ``the developer.''

\textbf{Stage 2: Constructs and Indicators --- specifying what $\mathbf{C}_{\text{after}} - \mathbf{C}_{\text{before}}$ means.}
Once stakeholders and goals have been specified, the framework next formalizes what meaningful capability change actually consists of for each group. For every goal identified in Stage~1, Stage~2 defines the underlying construct --- the real-world capability intended to improve --- and selects indicators that validly measure it. The resulting \textbf{Construct--Indicator Matrix} links stakeholder goals to measurable quantities and empirical instruments. 
This stage operationalizes the capability term in the utility equation and serves as the framework's primary defense against proxy displacement. Substitutions such as pass@k for software quality (S10), fluency for correctness (S6, S22), or institutional cost reduction for patient well-being (S16) must now be justified in terms of construct validity rather than benchmark convenience. The Stage~2 panel of Figure~\ref{fig:scu-walkthrough} therefore replaces generic benchmark metrics with deployment-relevant constructs including engineering velocity, skill acquisition, security efficacy, team collaboration, and end-user trust. Importantly, the figure illustrates that pass@k directly measures none of these constructs despite being the dominant benchmark metric used in the standard evaluation pipeline.

\textbf{Stage 3: Mechanisms of Change --- modeling how $\mathbf{C}$ evolves with $t$.}
Stage~3 introduces the temporal and causal structure underlying utility by identifying mechanisms through which AI interaction may improve or degrade capability trajectories: skill atrophy~\citep{BAINBRIDGE1983775}, seen in coding-assistant reliance reducing unassisted capability (S14); loss of productive struggle~\citep{Roediger2006-xp}; automation bias~\citep{Goddard2011-rp, Lyell2017-ma}; expertise reversal~\citep{Kalyuga01012003},  where initially low-performing users learn less than competent peers (S7); and failures of transfer. Each mechanism should yield explicit, testable predictions about how utility  evolves for particular users over time, providing the framework's central defense against temporal collapse: deployment-time performance is no longer conflated with durable capability. By specifying which longitudinal effects to expect, Stage~3 causally links the constructs of Stage~2 to the empirical protocols of Stage~4.

\textbf{Stage 4: Utility Measurement --- computing $\mathbf{U}$.}
With stakeholders, constructs, and expected dynamics now specified, Stage~4 should estimate utility directly through five protocols grounded in established human--AI interaction methodology \citep{Oulasvirta2026}.
\emph{Condition comparison} (human-alone, AI-alone, human+AI) should be able to isolate the AI's marginal contribution and tests for \emph{complementary team performance}---a benchmark rarely realized in practice~\citep{hemmer2024complementarityhumanaicollaborationconcept, 2021-ai-explanations-team-performance}. \emph{Longitudinal tracking} should measure the durability across horizons $t$. 
\emph{Subgroup disaggregation} should report $\mathbf{U}$ across Stage-1 stakeholder groups, since complementarity is strongly moderated by user expertise. \emph{Mechanism verification} tests whether the Stage-3 predictions (skill atrophy, automation bias, expertise reversal) manifest in observed interaction traces. 
% \emph{Procedural verification} should ensure auditability: every $(\text{prompt}, \text{suggestion}, \text{user action}, \text{outcome})$ tuple is written to a hash-chained, append-only trace store, and 
% Stage-3 mechanisms should be scored by pre-registered decoders (e.g., automation bias as the fraction of AI suggestions accepted without edit despite failing a held-out correctness check), so an independent party can replay the same decoders on the verified log and reproduce each $\mathbf{U}$ value within a declared tolerance. 
To support \emph{auditability}, interaction traces and mechanism scores can be logged in a reproducible format, allowing independent parties to replay the analysis and verify reported utility values.
The output of this stage would be a \textbf{Utility Profile} reporting $\mathbf{U}$ per $(u, G, t)$ tuple. In the DevAssist example (Figure~\ref{fig:scu-walkthrough}), this yields $\tilde{U} = +0.4$ for seniors, $\tilde{U} = -0.3$ for juniors, and $\tilde{U} = -0.5$ for security review---a profile aggregate benchmarks systematically obscure. 
Table~\ref{tab:stage4-protocols} summarizes how each of these protocols maps to the failure mode it prevents and the SCU-GenEval framework that supports it on DevAssist and  Epic Sepsis Model (S19). Together, the two columns show that the same Stage-4 machinery can be used either to plan a rollout or to diagnose why a deployed system failed to deliver utility.
\vspace{-1.2mm}
\vspace{-1.2mm}
\subsection{Three Instruments: Making Utility Measurement Affordable}
\label{subsec:instruments}
\vspace{-1.2mm}
We argue that the four stages are the right things to do during evaluation, but expensive on their own. Hence we propose three instruments which make them tractable. Each instrument plugs into specific stages and targets specific failure modes.

\textbf{Instrument 1: Structured Deployment Data Collection Protocol --- SDDCP (used in Stage 4; anchors Instruments 2 and 3).} We propose to have  SDDCP as a pre-registered, machine-readable contract completed \emph{before} deployment, specifying the setting (Stage 1), constructs and outcomes (Stage 2), expected mechanisms (Stage 3), comparison conditions and timeline (Stage 4), and subgroup disaggregation keys (Stages 1 and 4). It should be structured analogously to CONSORT-AI~\citep{Liu2020-zu} but adapted for socio-technical generative AI. The SDDCP should play three roles that can make SCU-GenEval deployable rather than aspirational: (i) pre-registration converts analyst degrees of freedom into auditable commitments, pre-empting post-hoc proxy substitution; (ii) it should serve as the ground-truth signal of $U$ in real settings, against which Instruments 2 and 3 should be calibrated;  and (iii) standardized SDDCP outputs contribute to a shared, public \textbf{AI Deployment Outcome Database} such that everyone can remain aware of the consequences before deployment.
%, enabling cross-organizational detection of site-generalization failures (S20-style) that no single deployer can observe alone. Domain-specific instantiations follow directly from the failure modes of Section~\ref{sec:gap}: mutation testing and CWE-incidence alongside pass@k for software (S10, S13); Expected Calibration Error and demographically-disaggregated reporting in healthcare~\citep{Obermeyer2019-na} (S16, S18, S22); unassisted-transfer assessment at four-week and six-month horizons in education~\citep{VanLEHN01102011} (S1, S3, S5). Had the SDDCP existed, S19 (Epic Sepsis) would have been caught pre-deployment.

\textbf{Instrument 2: Context-Conditioned User Simulators (used in Stages 3 and 4; calibrated against Instrument 1).} A simulator should instantiate a $(u, G, D)$ tuple as user behavior conditioned on persona attributes, goals, and contextual constraints --- e.g., ``junior developer under deadline pressure'' --- which can be realized through prompting, fine-tuning, or retrieval~\citep{bao2026eval4simevaluationframeworkpersona, tseng-etal-2024-two, abdulhai2025consistentlysimulatinghumanpersonas}, by grounding the simulator in measurable trait inventories such as Big Five dimensions inferred from behavioral logs to preserve statistical fidelity to real user distributions~\citep{PUB}, or by imposing structured action schemas (e.g., the Rating–Action–Response decomposition of RecUserSim~\citep{Chen_2025}) that enforce realistic decision structure rather than only plausible surface text.
For SCU-GenEval, we recommend that the simulators should be constructed from Stage-1 personas and Stage-3 mechanism predictions, then calibrated against SDDCP-collected ground truth (Instrument 1) before being used to estimate $\mathbf{U}$ at long $t$. They should estimate utility in days of compute rather than months of fieldwork, attacking temporal collapse, and refuse to model an ``average user,'' attacking distributional concealment. For DevAssist, a junior-engineer simulator could predict a 23\% drop in unassisted skill in nine days (Figure~\ref{fig:scu-walkthrough}, Instrument 2 panel) --- compressing what would otherwise be a year-long study.

\textbf{Instrument 3: Persona-Goal-Conditioned Proxy Metrics (used in Stage 2 for fast iteration; calibrated against Instrument 1).} PGCPMs are cheap metrics that approximate $\mathbf{U}$ by computing only on cases matching a specified $(u, G, D, t)$. Constructed in three steps, each grounded in an earlier stage: (i) pick the construct from Stage 2's matrix (defense against proxy displacement: the metric is tied to a validated construct); (ii) pick a validated measurement instrument~\citep{DeGrave2020-ee}; (iii) filter the evaluation set to cases matching the Stage 1 persona (defense against distributional concealment: refusing to average is the central design choice). PGCPMs are validated against the SDDCP ground truth from Instrument 1, again analogously to biomarker validation in clinical trials~\citep{Fleming2012-mj}. A junior-conditioned PGCPM would have caught the DevAssist regression three model versions earlier than aggregate pass@k (Figure~\ref{fig:scu-walkthrough}).

\vspace{-2.6mm}
\section{Calls to Action for Diverse Domains}
\label{sec:calls-to-action}
\vspace{-2.2mm}
The benchmark--utility gap will not disappear on its own.
Each domain requires concrete commitments from researchers, funders, regulators, and deploying organizations to operationalize utility measurement. Hence, we translate the framework of Section~\ref{sec:proposal} into domain-specific calls to action, grounded in the failure modes documented in Section~\ref{sec:failure-modes}.

\vspace{-2mm}
\paragraph{Education.}
The biggest problem in education is \textbf{temporal collapse}: students may perform better while using AI, but this does not always mean they actually learn or retain knowledge long-term (S1, S3, S4, S5). We call on:
(i) educational AI companies to report whether students can perform tasks \emph{without AI help} after weeks or months, not just during immediate use;
(ii) AI-in-education conferences and journals to report results separately for students from different backgrounds and skill levels, since average scores can hide inequality (S7); and
(iii) funding agencies such as NSF and IES to require long-term deployment studies before supporting classroom AI systems.

\vspace{-2mm}
\paragraph{Software engineering.}
The main problem in coding AI is \textbf{proxy displacement}: metrics like pass@k often look good even when generated code is insecure or hard to maintain (S10, S13, S15). Another issue is that AI tools affect junior and senior developers differently (S14). We call on:
(i) benchmark creators to report security and robustness measures alongside pass@k;
(ii) coding-assistant companies to publish results separately for junior, mid-level, and senior developers (Instrument~1); and
(iii) software-security organizations such as OWASP and NIST to include user-specific safety evaluations in software assurance standards.

\vspace{-2mm}
\paragraph{Healthcare.}
The biggest issue in healthcare AI is \textbf{distributional concealment}: systems may work well on average but fail for certain demographic groups or hospitals (S16, S17, S18, S20). We call on:
(i) regulators such as the FDA and EMA to require testing on real external patient populations before approval;
(ii) medical journals to report calibration and subgroup performance, not only AUC scores; and
(iii) hospitals and healthcare systems to share deployment outcome data publicly so failures can be identified across institutions instead of being repeatedly rediscovered (Instrument~1).
\vspace{-2.2mm}

\paragraph{Law, civil rights, and adjudication.} The main problem in legal and civil-rights AI is \textbf{proxy displacement}: systems often replace legally meaningful reasoning with fluent answers, proprietary scores, or automated recommendations that affected people cannot easily verify or contest. COMPAS showed this pattern in criminal sentencing (S25); generative legal tools repeat it when fluent answers contain fabricated or unverifiable citations (S26, S28);  and tenant-screening tools can hide protected-class disparities behind automated eligibility scores (S27). We call on: (i) bar associations and courts to require independent, preregistered hallucination and citation audits for legal AI, reported by jurisdiction, role, and query type (Instrument~1); and (ii) civil-rights agencies such as DOJ and HUD to require protected-class-disaggregated evaluations of automated screening and eligibility tools  (Instrument~3), so S27-style disparities surface during deployment rather than through litigation.

\paragraph{Cross-domain infrastructure.} Beyond domain-specific calls, we call on the AI research community to (i) establish a standing Best Deployment Study award, mandate pre-registration for any paper claiming real-world utility (in A* conferences like NeurIPS, ICLR, CVPR, CHI), and adopt Registered-Report-style multi-cycle review so that longitudinal, stakeholder-grounded evaluation becomes a path to recognition; (ii) maintain a federated AI Deployment Outcome Repository with public deployment metadata and access-controlled outcome data, enabling simulator calibration (Instrument~2) and proxy-metric validation (Instrument~3) across organizations without exposing raw user data;
% maintain a shared, open AI Deployment Outcome Database that calibrates simulators (Instrument~2) and validates proxy metrics (Instrument~3) across organizations;
and (iii) establish reporting standards analogous to CONSORT and TRIPOD, but adapted for socio-technical AI deployment, so that utility profiles become the unit of comparison.

% =============================================================
% Section 5: Alternative Views
% =============================================================
\vspace{-2.6mm}
\section{Alternative Views}
\label{sec:alternative-views}
\vspace{-1.2mm}
\paragraph{Better benchmarks are sufficient.} A natural objection is that the failures in Section~\ref{sec:failure-modes} reflect \emph{bad} benchmarks rather than the \emph{category} of benchmarks, and that contamination-resistant, dynamic, or harder evaluations would close the gap. We partially agree: harder benchmarks are valuable, and recent dynamic evaluations \citep{zhu2024dyvaldynamicevaluationlarge, livebench} are genuine progress. However, this view does not address construct invalidity (S6, S16, S22): no amount of benchmark difficulty converts cost into health need or fluency into correctness. It also does not address temporal collapse (S3, S5, S14), since static benchmarks measure no trajectory however hard they are. SCU-GenEval is complementary to better benchmarks, not opposed to them.
\vspace{-1.5mm}
\paragraph{Utility evaluation is too expensive to scale.} A pragmatic objection is that longitudinal, disaggregated, deployment-grounded measurement is incompatible with the iteration speed of modern model development. 
We grant the cost concern---it is precisely why we propose the three instruments of Section~\ref{sec:proposal}. 
User simulators (Instrument~2) compress longitudinal effects into days of compute; PGCPMs (Instrument~3) provide cheap, validated proxies for everyday development; the SDDCP (Instrument~1) is amortized across the deployment lifecycle. Moreover, the cost framing is asymmetric: the cases in Table~\ref{tab:cases} each carry deployment costs---patient harm, security breaches, learning loss, eroded trust---that dwarf the cost of pre-deployment utility evaluation. Relevant comparison is not benchmark cost vs.\ utility-evaluation cost, but utility-evaluation cost vs.\ \emph{the cost of harms it prevents}.
\vspace{-1.5mm}
\paragraph{Stakeholder-centered evaluation is too subjective.} Another objection is that naming stakeholders, constructs, and mechanisms (Stages~1--3) introduces analyst degrees of freedom that undermine reproducibility. We respond that construct validity is a mature methodological tradition \citep{Cronbach1955-gd,Jacobs_2021}, not a novel source of subjectivity, and that the alternative---implicit construct selection by benchmark authors---is \emph{more} subjective, not less, because it is unstated. Pre-registration of Stakeholder--Goal Map and Construct--Indicator Matrix (as required by SDDCP) makes the analyst's choices auditable in a way that opaque benchmark choices are not.
\vspace{-1.2mm}
\paragraph{Post-deployment monitoring already addresses this.} A further objection is that real-world harms are best caught by deployment monitoring (e.g., pharmacovigilance-style surveillance) than by pre-deployment utility evaluation. 
We agree monitoring is essential, but insufficient: it detects harms after rollout rather than preventing harmful deployments. Healthcare shows the limitation: surveillance may reveal clinical AI failures, but not whether they stem from the model, clinician--AI interaction, or local workflow.
% We agree monitoring is essential, but it is not sufficient. Monitoring detects harms; it does not prevent the deployments that cause them. Healthcare illustrates the limitation: after clinical AI enters workflow, surveillance may reveal poor calibration, alert burden, or missed cases, but not whether the cause was the model, the clinician--AI interaction, or the local workflow.
SCU-GenEval is thus designed to operate before and during deployment, with the SDDCP providing the bridge to post-deployment surveillance.
\vspace{-1.2mm}
\paragraph{HCI and human-factors research already do this.} Finally, one might argue that the human-factors and HCI communities have long evaluated socio-technical systems against human outcomes, and that AI evaluation should simply adopt those methods. We strongly endorse the underlying methodological tradition. Our contribution is not to invent stakeholder-centered evaluation but to argue that \emph{the AI benchmarking community has not adopted it}, and to provide a framework specifically adapted to generative AI's distinctive failure modes (LLM-as-judge circularity, contamination, scale-induced distributional concealment) that classical human-factors methods were not designed to handle.

\paragraph{Simulators reintroduce the problem they claim to solve.} \emph{A self-undermining objection targets Instrument~2: if construct validity is the central concern, simulating users with LLMs is suspect, since the simulator may inherit the same biases and blind spots as the system under test.} This is the sharpest objection and the one we take most seriously.
In SCU-GenEval, simulators are not treated as ground truth and are not allowed to certify deployment utility on their own. They must first be calibrated against SDDCP-collected deployment data from Instrument~1 ~\citep{Fleming2012-mj}. Their role is narrower: to stress-test plausible temporal trajectories, such as skill atrophy, expertise reversal, overreliance, or failures of transfer, before these harms become visible in full deployment. 
% Simulators are therefore early-warning and scaling tools, while Instrument~1 remains the evidentiary anchor for utility claims.
% We would like to argue in the following manner: 1) the simulators in SCU-GenEval are calibrated against SDDCP-collected ground truth (Instrument~1) before we begin to trust them ~\citep{Fleming2012-mj}.
% 2) We propose that the purpose of the simulators is to deal with temporal—skill atrophy, expertise reversal, durable transfer where they do not specifically need to replicate users. They should be good enough to highlight the  trajectories that might cause deterioriation of utility. 

We admit that some of the researchers will treat these counterarguments differently. 
Therefore, we highlight this argument that the field is unlikely to abandon benchmarks, nor should it. 
Infact, the current benchmarks hold immense value for the low-stakes capability tracking and rapid iteration.
Our claim is: \emph{for any system whose deployment causes non-trivial consequences, benchmark performance is necessary but not sufficient evidence of utility, and the field currently behaves as if it were sufficient.}

\vspace{-2.6mm}
\section{Conclusion}
\vspace{-2.6mm}
In deployment scenarios, benchmark performance alone is no longer sufficient to measure real-world usefulness (justified through 28 case studies). 
Our proposed framework, SCU-GenEval, calls for a shift in the evaluation mechanism from artifact quality to human outcome trajectories, while the proposed instruments make stakeholder-conditioned utility evaluation practical. 
We do not claim utility evaluation is easy; we claim it is overdue. 
For systems with meaningful human consequences, utility measurement should necessarily become a first-class research objective alongside benchmarks.
% %\jbgcomment{I think that previous sections run a little longer than they need to and this section should present a clearer plan for how to actually implement your agenda}

% %Broader Implications: For AI safety (testing manipulation), human-AI collaboration, and creativity.
% We have argued that Language Games offer an ideal and necessary evaluation framework for studying AI, particularly as models display emergent abilities.
% This stance is based on the fundamental limitations of automatic and human evaluations.
% Even if researchers address these limitations, inherent flaws in the evaluation designs prevent them from accurately and comprehensively measuring AI capabilities.
% While current benchmarks are sufficient for evaluating state-of-the-art LLMs, they will become obsolete as AI evolves.
% Better evaluation methods will become increasingly critical as AI progresses.
% Language Games provide a robust solution to this ongoing evaluation challenge.

\bibliographystyle{plainnat}
\bibliography{bib/journal-full,bib/jbg}

\section*{Appendix}

{\renewcommand{\arraystretch}{1.15}
\begin{longtable}{@{}p{0.6cm}p{3.2cm}p{7.5cm}p{2.0cm}@{}}
\caption{Twenty-four documented cases of benchmark--utility divergence. Each system
scored well on a benchmark designed for tractability while failing on the
construct that mattered in deployment.}
\label{tab:cases}\\
\toprule
\textbf{ID} & \textbf{Benchmark metric} & \textbf{Utility gap $\rightarrow$ real-world cost} & \textbf{Evidence}\\
\midrule
\endfirsthead
\multicolumn{4}{@{}l}{\textit{Table~\ref{tab:cases} continued from previous page}}\\
\toprule
\textbf{ID} & \textbf{Benchmark metric} & \textbf{Utility gap $\rightarrow$ real-world cost} & \textbf{Evidence}\\
\midrule
\endhead
\midrule
\multicolumn{4}{r@{}}{\textit{continued on next page}}\\
\endfoot
\bottomrule
\endlastfoot
\multicolumn{4}{@{}l}{\cellcolor{eduFill!50}\textit{\textbf{Education}}}\\
S1 & In-system accuracy, completion & ITS achieves gains comparable to human tutoring on practiced problems, but distal/transfer outcomes are infrequently measured & \cite{VanLEHN01102011} \\
S2 & Vocabulary, sentence complexity, length & ETS e-rater consistently awarded BABEL-generated essays scores, despite the essays being semantically incoherent & \cite{Perelman}\\
S3 & Assignment completion, submission quality & Post-ChatGPT, take-home completion rates rise while concerns persist that gains do not transfer to unassisted, in-class assessment & \cite{Cotton03032024}\\
S4 & Immediate comprehension, recall & Retrieval-practice research establishes that effortful retrieval drives long-term retention; AI-generated summaries and elaborated notes plausibly degrade this mechanism by removing the retrieval step & \citep{Roediger2006-xp, Retrieval}\\
S5 & Autograder scores, completion & CS1 students using Copilot were observed to experience metacognitive difficulties, including \emph{shepherding} (modifying code to align with model suggestions) and \emph{straying} (following incorrect suggestions into prolonged debugging dead ends) & \cite{Prather2023}\\
S6 & Fluency, coherence, perceived helpfulness & LLM-generated scholarly content included fabricated citations, incorrect author attributions, and nonexistent references despite appearing fluent and convincing to readers & \cite{Walters2023-rf}\\
S7 & Aggregate test-score improvement & Personalized learning systems often produce heterogeneous outcomes: students with lower prior performance or reduced working-memory capacity may benefit substantially less than stronger peers, allowing aggregate gains to mask widening educational inequities & \cite{Dumont2023-jc} \\
S8 & Grammar, rubric adherence & Co-writing with feedback-tuned LLMs such as InstructGPT was associated with a statistically significant reduction in lexical and content diversity across authors, suggesting convergence toward homogenized writing styles; this effect was not significant for base GPT-3 & \cite{padmakumar2024doeswritinglanguagemodels}\\
S9 & Factual accuracy, topic coverage & LLM-generated medical multiple-choice questions disproportionately concentrated on lower Bloom’s taxonomy levels such as recall and comprehension, while underrepresenting higher-order reasoning skills including application and analysis & \cite{Law2025} \\
\addlinespace
\multicolumn{4}{@{}l}{\cellcolor{swFill!50}\textit{\textbf{Software engineering}}}\\
S10 & pass@k, unit-test success & $\sim$40\% of Copilot completions contained a CWE-listed vulnerability (e.g., SQLi, buffer overflow), independent of pass-rate metrics & \cite{Asleep}\\
S11 & Plausible-patch rate, time-to-patch & LLM-based program-repair frequently produces patches that pass the available test suite while leaving the underlying defect unresolved (the long-standing plausible-vs-correct gap in APR) & \cite{dakhel2023githubcopilotaipair} \\
S12 & Line-level defect recall & Static AI code review captures pattern-matchable implementation bugs but systematically misses design-level defects (architectural boundary violations, missing access checks)---roughly half of security defects per McGraw's classic split & \citep{mcgraw2006software, charoenwet2024empiricalstudystaticanalysis} \\
S13 & Line and branch coverage & Code coverage substantially overstates the fault-detection power of LLM-generated tests & \cite{Wang_2026}\\
S14 & AI-assisted productivity, tickets closed & Heavy reliance on AI coding assistants is associated with reduced unassisted problem-solving capability, consistent with the classical \emph{ironies of automation} and skill-decay literature & \citep{Prather2024, BAINBRIDGE1983775, Parasuraman2010-gq}\\
S15 & BLEU, textual similarity & Generated API documentation can score well on textual-similarity metrics while containing factual errors and not correlating with human & \citep{KhanUddin, Correlating}\\
\addlinespace
\multicolumn{4}{@{}l}{\cellcolor{hcFill!50}\textit{\textbf{Healthcare}}}\\
S16 & AUC for healthcare cost prediction & Cost used as a proxy for need: at equal risk score, Black patients had 26.3\% more chronic illnesses than White patients; correcting the bias would raise Black auto-enrolment from 17.7\% to 46.5\% (a $\sim$28.8\,pp gap). The broader class of cost-prediction algorithms affects $>$100M U.S.\ patients & \cite{Obermeyer2019-na}\\
S17 & Classification AUC & Deep-learning COVID-19 chest-radiograph classifiers learn confounders (data-source/site, scanner artefacts, patient positioning) rather than pulmonary pathology, and fail under distribution shift & \citep{DeGrave2020-ee, OAKDENRAYNER2020106}\\
S18 & Response quality, benchmark accuracy & GPT-4 shows some discrimination against certain race/gender, and hence the assessment became biased & \cite{Zack2024-zn}\\
S19 & Internal proprietary validation & External retrospective validation of Epic Sepsis Model showed high or reasonable AUC performance & \citep{Wong2021-fi, Habib-EpicSepsis2021}\\
S20 & Training-site concordance & IBM documents (internal) revealed that Watson for Oncology which was developed using small amount of synthetic cases and treatment preferences from a single institution produced recommendations that clinicians reportedly deemed “unsafe and incorrect.” & Ross \& Swetlitz, STAT (2018) \footnote{\url{https://www.statnews.com/2018/07/25/ibm-watson-recommended-unsafe-incorrect-treatments/}}\\
S21 & ROUGE, perceived usefulness & Automation bias is a valid concern in clinical decision support, where clinicians may place undue trust in AI-generated recommendations without sufficiently validating the underlying evidence or patient data & \citep{Goddard2011-rp, Lyell2017-ma}\\
S22 & Fluency, medical QA accuracy & Med-HALT and other such medical hallucination benchmarks demonstrate that medical LLMs can generate plausible yet fabricated lab values, medication names, and clinical facts with high apparent confidence, such failures frequently evade detection by fluency-based or surface-level QA evaluation metrics & \cite{Bhayana2023-iw}\\
S23 & Sensitivity, alert detection & A large-scale ICU study reported that a substantial fraction of annotated arrhythmia alarms were false positives, contributing to alarm fatigue—a well-documented factor associated with missed or delayed responses to critical clinical events & \citep{Drew}\\
S24 & Engagement, short-term PHQ-9, satisfaction & Mental-health chatbots show short-term symptom reductions (e.g., Woebot at 2--3 weeks); systematic reviews report weak and heterogeneous evidence for clinically meaningful benefit at longer follow-up & \citep{Fitzpatrick2017-tr, Abd-Alrazaq2020-bj}\\

\multicolumn{4}{@{}l}{\cellcolor{lawFill!50}\textit{\textbf{Law, civil rights, and adjudication}}}\\
S25 & AUC for two-year recidivism  & COMPAS scores achieve acceptable AUC but are unverifiable by defendants because the algorithm's weights are 
proprietary; in \emph{State v. Loomis}, the defendant could not contest how sex factored into his ``high risk'' score, yet the sentence was upheld with only a 
written advisement to judges. & \citep{state-vs-loomis-2016, Rudin-compas-2020}\\
S26 & Fluency, surface citation form, Self-reported accuracy & ChatGPT generated six fully fabricated judicial opinions with plausible formatting; attorneys relied on the model's self-confirmation that cases ``exist''; falsely named judges notified by court order and sanctions imposed. &
% Procedural norm of Rule 11 attorney verification was unverifiable from model output alone. 
% MATA v. AVIANCA INC (2023) \footnote{\url{https://caselaw.findlaw.com/court/us-dis-crt-sd-new-yor/2335142.html}}
\citep{mata-avianca-ChatGPT2023}
\\
S27 & SafeRent Score (proprietary tenant-screening score on a 200-800 scale) &
SafeRent's algorithmic tenant-screening score allegedly failed to account for the financial protection provided by housing vouchers, which cover about 73\% of monthly rent on average, and produced disproportionately lower scores for Black and Hispanic applicants. SafeRent denied wrongdoing but agreed for five years to stop issuing automated ``approve/decline'' recommendations for voucher applicants in covered circumstances. & \cite{safeRent-rentDecline-2023} \\
S28 & Fluency, Hallucination rate &
Legal AI tools hallucinated despite vendor claims: Westlaw AI-Assisted Research hallucinated in 33\% of legal queries; so faster research still required lawyers to verify authority existence, validity, jurisdiction, and support & \cite{hallucinate-legalAI-2025}\\
\end{longtable}
}

\rowcolors{2}{white}{rowgray}

\begin{table}[h]
\centering
\small
\caption{The five Stage-4 measurement protocols, the framework components each protocol draws on, and how each applies in two modes: prospective rollout (DevAssist, Figure~\ref{fig:scu-walkthrough}) and retrospective diagnosis (Epic Sepsis Model, S19~\citep{Wong2021-fi, Habib-EpicSepsis2021}).}
\vspace{0.2cm}
\label{tab:stage4-protocols}
\begin{tabular}{@{}p{1.6cm}p{3.4cm}p{2.0cm}p{2.6cm}p{3.0cm}@{}}
\toprule
\textbf{Protocol} &
\textbf{What it does $\rightarrow$ Failure prevented} &
\textbf{Framework support} &
\textbf{DevAssist (prospective)} &
\textbf{Epic Sepsis Model (retrospective)} \\
\midrule
\textbf{Condition comparison} &
Compares human-alone, AI-alone, and human+AI performance to estimate the AI's marginal contribution $\rightarrow$ \textbf{prevents proxy displacement} by avoiding crediting the AI for gains produced by the human or workflow. &
Stage 2 indicators; SDDCP comparison conditions &
Shows that assisted productivity gains aren't uniform: seniors gain, juniors may lose unassisted capability. &
Would reconstruct clinician-alone, model-alone, and clinician+alert conditions; e.g., whether ESM alerts identify sepsis cases not already receiving antibiotics~\citep{Wong2021-fi}. \\
\addlinespace
\textbf{Longitudinal tracking} &
Measures utility across multiple horizons, including unassisted performance after AI exposure $\rightarrow$ \textbf{prevents temporal collapse} by catching delayed harms (skill decay, overreliance) that snapshot metrics miss. &
Stage 3 mechanisms; SDDCP timeline; calibrated simulators &
Reveals whether short-term productivity for juniors masks later unassisted skill loss. &
Would track ESM utility across deployment horizons rather than assuming pre-deployment validation persists. \\
\addlinespace
\textbf{Subgroup disaggregation} &
Reports utility per stakeholder--goal--context tuple rather than averaging across users or settings $\rightarrow$ \textbf{prevents distributional concealment} by exposing subgroup- or context-specific harms. &
Stage 1 personas; Stage 2 indicators; SDDCP keys; PGCPMs &
Reports $\tilde{U}=+0.4$ (seniors), $-0.3$ (juniors), $-0.5$ (security review), not a misleading positive aggregate. &
Would stratify ESM performance by ward, cohort, site, and workflow rather than relying on aggregate AUC. \\
\addlinespace
\textbf{Mechanism verification} &
Tests whether mechanisms predicted in Stage 3 appear in logs and behavior traces $\rightarrow$ \textbf{\textit{prevents mechanism misspecification}} by checking whether the assumed pathway of benefit or harm is real. &
Stage 3 mechanisms; SDDCP traces; simulators &
Checks logs for shepherding, straying, skill atrophy, automation bias, expertise reversal. &
Would use alert timestamps, clinician actions, and override patterns to test for ignored alerts and alert fatigue. \\
\addlinespace
\textbf{Auditability} &
Records settings, conditions, timelines, subgroup keys, and analysis decisions in a reproducible format $\rightarrow$ \textbf{\textit{prevents post-hoc proxy substitution}} by allowing outside parties to recompute the utility profile. &
SDDCP; pre-registration; replayable logs; outcome database &
Logs every suggestion-action-outcome tuple so utility can be independently recomputed. &
Would make cohort definition, sepsis labels, thresholds, alert counts, and timing comparisons replayable for external recomputation. \\
\bottomrule
\end{tabular}
\end{table}

\section{From Prospective Evaluation to Retrospective Diagnosis.}
\label{app:table2-discuss}

Table~\ref{tab:stage4-protocols} extends the Stage~4 discussion from 
Section~\ref{subsec:framework} by showing that the same protocols apply 
in two evaluation modes. In the DevAssist example 
(Figure~\ref{fig:scu-walkthrough}), the protocols are \emph{prospective}: 
they specify what evidence should be collected before rollout. In the 
Epic Sepsis Model (ESM) case (S19), the protocols are 
\emph{retrospective}: they organize what evidence is needed to diagnose 
why a deployed clinical model did or did not produce 
utility~\citep{Habib-EpicSepsis2021, Wong2021-fi}.

This reframes the ESM evaluation question. The original framing was 
``what was the model's AUC?''; the protocols generate a different 
sequence of questions: 
\begin{itemize}
    \item \textit{Condition comparison}: Did the alert add value beyond existing clinician practice?
    \item \textit{Longitudinal tracking}: Did performance remain stable across time and cohorts?
    \item \textit{Subgroup disaggregation}: Were failures concentrated in particular settings or workflows?
    \item \textit{Mechanism verification}: Did clinicians act on, ignore, or over-rely on alerts?
    \item \textit{Auditability}: Could external researchers replay the evidence behind the utility claim?
\end{itemize}

Retrospective diagnosis cannot be reduced to recomputing predictive 
performance. It requires recovering the comparison condition, time 
horizon, affected subgroups, behavioral mechanism, and audit trail 
underlying the original utility claim --- the five Stage~4 protocols 
defined in Section~\ref{subsec:framework}. Table~\ref{tab:stage4-protocols} 
is therefore not a summary of ESM but a diagnostic schema: a translation 
from retrospective case evidence into auditable utility claims via the 
Stage~4 protocols and Section~\ref{subsec:instruments} instruments.

\end{document}